# ReMIND: Orchestrating Modular Large Language Models for Controllable Serendipity
## A REM-Inspired System Design for Emergent Creative Ideation

Makoto Sato

Mathematical Neuroscience Unit, Institute for Frontier Science Initiative, Laboratory of Developmental Neurobiology, Graduate School of Medical Sciences, Kanazawa University, Kanazawa, Ishikawa, Japan.

Correspondence: makotos@staff.kanazawa-u.ac.jp

**Abstract**

Large language models (LLMs) are increasingly used not only for problem solving but also for creative ideation; however, generating ideas that are both novel and coherent remains challenging. While high-temperature sampling can promote originality, it often compromises consistency and usefulness. Here, we propose ReMIND, a four-stage framework comprising **wake**, which establishes a stable semantic baseline through low-temperature generation; **dream**, which performs high-temperature exploratory generation; **judge**, which evaluates candidate outputs for consistency and extracts salient novel ideas; and **rewake**, which consolidates selected ideas into coherent final outputs. By assigning these functions to independent LLM modules, ReMIND explicitly separates exploration from stabilization. We systematically evaluated diverse model configurations using multiple creative ideation tasks. External evaluations showed that novelty enhancement could emerge through two separable stages: first from the wake to the dream phase through high-temperature exploration, and subsequently from the dream to the rewake phase through judge-mediated selection and consolidation. Notably, different LLM families exhibited distinct functional tendencies. These findings suggest that serendipitous ideation in LLMs is not solely a property of individual models but emerges from interactions among models assigned to specialized cognitive roles. ReMIND provides a general framework for investigating computational creativity and demonstrates how modular LLM orchestration can bridge exploratory generation and coherent idea formation.

## Introduction

Large language models (LLMs) have rapidly evolved from tools for language understanding and task execution into systems capable of producing creative outputs, including stories, hypotheses, metaphors, and other forms of ideation. Recent studies have shown that LLMs can perform well on divergent-thinking tasks and generate research ideas judged to be highly novel, although feasibility and consistency remain variable [1,2]. This expansion has prompted growing interest in computational creativity: how artificial systems can generate ideas that are not only fluent, but also genuinely novel and meaningful, thereby supporting human creative thinking.

In human cognition, however, creative insight often emerges not during deliberate, goal-directed thought, but during offline or loosely constrained cognitive states. Psychological and neuroscientific studies have long associated such insights with sleep—particularly rapid eye movement (REM) sleep—during which memory traces are reorganized, associations are loosened, and unusual conceptual combinations arise [3,4]. Importantly, these processes are not purely random. Empirical and theoretical work in cognitive science suggests that unconstrained exploration is typically followed by phases of stabilization, evaluation, and reintegration into coherent waking cognition [5].

While the biological mechanisms underlying REM sleep remain complex, its functional organization suggests a general computational principle: creativity may benefit from the temporary separation of associative exploration from executive evaluation. During REM sleep, prefrontal regions involved in evaluative and executive control show reduced activity, while internally generated associative processing remains prominent [6]. This perspective is consistent with classical theories of creativity, which emphasize remote association and the recombination of disparate concepts as sources of novel ideas [7,8].

Existing approaches to creative LLM generation have used iterative self-feedback and refinement, or natural-language feedback from previous attempts to guide subsequent responses [9,10]. Other frameworks coordinate multiple role-defined agents through dialogue, critique, and collaborative ideation [11-13]. However, these approaches are primarily designed to optimize output quality rather than to isolate the functional contributions of exploration, evaluation, and consolidation. This leaves open the question of whether these processes can be separated and studied as distinct stages of creative generation.

Here, we propose ReMIND, a REM-inspired framework for serendipitous idea generation that explicitly separates exploration, evaluation, and consolidation into distinct functional modules (Figure 1). Rather than relying on a single monolithic generation process, ReMIND orchestrates multiple LLM instances with complementary roles: a **wake** module that produces a stable, low-temperature baseline response; a **dream** module that performs high-temperature stochastic generation to explore unconventional conceptual combinations beyond that baseline; and a **judge** module that evaluates candidate outputs for consistency and extracts salient novel ideas. Selected candidate ideas are then passed back to the wake module in a consolidation step—**rewake**—which reformulates promising ideas into a coherent final output.

Crucially, in ReMIND the initial wake output serves as a baseline rather than a direct contributor to the final text. The creative transformation occurs through the dream–judge–rewake loop, allowing the system to depart substantially from the original response while retaining relevance and internal structure. This design mirrors a functional interpretation of REM sleep, in which exploratory generation is transiently decoupled from evaluative control, and coherence is restored only after selective consolidation.

In this analogy, exploratory generation corresponds functionally to the reactivation and recombination of memory representations during sleep [4], while reduced evaluative control parallels the attenuation of prefrontal activity during REM sleep [6]. The subsequent restoration of coherence in the rewake phase may be functionally analogous to the integration of learned knowledge with inferences formed through sleep-dependent processing [5,14,15].

To assess whether ReMIND produces ideas that are meaningfully distinct from baseline prompt-following behavior, we conduct systematic sweeps across prompts, output word limits, temperatures, and random seeds. We evaluate outputs using two complementary approaches: (i) embedding-based similarity metrics that quantify semantic distance between baseline and high-temperature outputs, and (ii) independent, external LLM-based evaluations that score novelty. This dual evaluation strategy allows us to distinguish trivial variation from genuine conceptual transformation.

Our results show that ReMIND can produce outputs that are both novel and coherent, although high-quality ideas emerge only in a subset of runs, highlighting the inherently probabilistic nature of serendipitous creativity. This observation motivates viewing creative generation not as a deterministic capability of a model, but as a process that benefits from structured exploration, selective filtering, and consolidation. Thus, we introduce ReMIND as a modular, REM-inspired framework for controllable serendipitous idea generation in LLMs. We demonstrate that modular orchestration offers a practical path toward studying and engineering creative emergence in artificial intelligence.

**Methods**

## 1. Overview of the ReMIND framework

ReMIND (REM-inspired Modular Ideation Network for Discovery) is a modular framework for serendipitous idea generation in LLMs, inspired by functional interpretations of REM sleep. The central design principle of ReMIND is the explicit separation of exploration, evaluation, and stabilization into distinct computational stages, each implemented by a dedicated LLM module orchestrated through Python-based pipelines. In our experiments, these modules were instantiated using locally hosted LLMs accessed via the llama_cpp interface, enabling fine-grained control over sampling parameters, random seeds, and execution order.

Information in ReMIND flows through the following sequence (Figure 1):

**Prompt → wake**

**→ dream → judge → rewake**

The **wake module** produces a low-temperature, high-consistency baseline response to the input prompt (*wake_out*). The **dream module** then performs high-temperature stochastic sampling, raising the probability of otherwise low-likelihood tokens and thereby producing outputs that deviate from the model's typical response (*dream_out)*. Outputs from the wake and dream modules are assessed by an independent **judge module**, which evaluates outputs for consistency and extracts a salient novel idea, when present. Here, *idea_wake* and *idea_dream* denote the core concepts extracted by the judge from the wake and dream outputs, respectively. Candidate ideas (*idea_dream*) are subsequently passed to a consolidation step termed **rewake**, in which the wake model is reused to re-articulate selected ideas into a coherent final output.

Although the wake model appears twice in this pipeline, its functional role differs fundamentally between stages. In the initial wake phase, it provides a stable semantic anchor, whereas in the rewake phase it serves as a stabilizing and compressive mechanism that integrates exploratory content into an interpretable representation.

## 2. Wake module (baseline generation)

The **wake module** provides a structured baseline response to a given prompt. It operates at temperature 0 to maximize internal consistency, relevance, and semantic alignment with the input. This module establishes a stable reference point that reflects the model's conventional, high-confidence interpretation of the prompt.

## 3. Dream module (exploratory generation)

The **dream module** is responsible for exploratory idea generation. It operates at elevated temperatures (1, 3, 10), increasing the probability of less likely outputs and promoting unconventional conceptual combinations, or at low temperatures (0, 0.3, 0.6) as controls. This encourages analogical leaps, and semantic drift beyond the wake response.

### *4.* Judge module (independent evaluation)

In the formal ReMIND pipeline, only dream outputs (*dream_out*) are processed by the judge for consistency assessment and novel-idea extraction prior to rewake. For analytical control, wake outputs (*wake_out*) were also processed by the same judge to extract corresponding core concepts. Here, *idea_wake* and *idea_dream* denote the concepts extracted from the wake and dream outputs, respectively. Only *idea_dream* is passed to the rewake stage, in which the wake model re-articulates the selected concept into a coherent final output. The judge does not participate in generation; it operates as a separate LLM instance at a deterministic setting (temperature = 0) and is instantiated independently from the wake and dream modules, even when the same underlying model family is used.

### 5. Rewake (consolidation and articulation)

Selected candidate ideas are passed to a final **rewake** step. In this stage, the original wake model is reused to re-articulate the selected idea in a coherent, structured form. The rewake stage is not explicitly prompted to generate additional novelty; rather, it is designed to stabilize and articulate the selected exploratory content. Notably, the final output is not a direct continuation of the initial wake response. Instead, it reflects a transformation in which exploratory dream content (*idea_dream*)—filtered by the judge—is re-expressed through low-temperature generation by the wake model (temperature = 0).

### 6. Review phase: quantitative evaluation of ideation quality

LLM-based evaluators have been used as scalable approaches for assessing texts according to predefined evaluation criteria [16,17]. To evaluate ideation quality in a systematic and reproducible manner, the output of ReMIND was passed through the **review** phase implemented using an external LLM (gpt-5.2 in this study). Outputs from the wake, dream, and rewake stages (*wake_out, dream_out, rewake_out*) were evaluated independently along three predefined dimensions:

- **Alignment**: relevance and responsiveness to the original prompt
- **Coherence**: internal logical consistency, clarity, and structural integrity
- **Novelty**: degree of conceptual departure from conventional or expected interpretations

  Scores were assigned on a discrete five-point Likert scale (1–5) for each dimension.

## 7. **Embedding-based similarity analysis**

To quantify semantic transformation across the ReMIND pipeline, we performed an embedding-based similarity analysis between generated outputs. Cosine similarity was computed between the full-text outputs of the wake module (*wake_out*) and the dream module (*dream_out*). Outputs were encoded using the pretrained SentenceTransformer model, all-MiniLM-L6-v2 [18]. This metric captures the degree of semantic overlap while remaining relatively insensitive to surface-level lexical variation. As a negative control, cosine similarity was computed between wake outputs (*wake_out*) generated under identical conditions. Cosine similarity values close to 1 indicate strong semantic preservation relative to the baseline response, whereas lower values indicate substantial conceptual divergence.

## 8. **Parameter sweeps and experimental design**

To systematically investigate the conditions under which ReMIND produces serendipitous ideas, we conducted extensive parameter sweeps across prompt configurations, sampling temperatures, output word limits, and random seeds. For each parameter combination, a single ReMIND execution was performed, yielding a baseline wake output (*wake_out* and *idea_wake*), an exploratory dream output (*dream_out* and *idea_dream*), and a consolidated rewake output (*rewake_out*), which were subsequently analyzed using external LLM-based review and embedding-based similarity measures.

## 9. **Prompt design**

### Wake and dream prompts

The wake and dream modules share the same base prompt templates, designed to elicit conceptual interaction between two separate ideas. These prompts explicitly reference two concepts (X and Y) and request comparative interpretation or ideation within a fixed word limit (Z), but do not bias the model toward specific rhetorical forms. The functional distinction between wake and dream arises not from prompt content, but from sampling configuration. The wake module operates at low temperature to produce a stable, high-consistency response, whereas the dream module uses elevated temperature to promote exploratory and unconventional generation.

Examples of the concept pairs included:

a. Physical concepts (*time* and *space*; Time & Space)
b. Chemical classification systems and symbolic interpretation frameworks (*periodic table* and *tarot divination*; Periodic Tarot)
c. Mathematical structures and cultural practices (*aperiodic tiling* and *traditional craft*; Aperiodic Craft)

Prompts were formulated in a neutral, descriptive style to elicit conceptual interaction between two domains. Specifically, we employed three prompt templates:

1. "Compare the meaning of X and Y within Z words."
2. "Describe the unexpected relationship between X and Y within Z words."
3. "Propose a new idea about the relationship between X and Y within Z words."

These templates differ in their degree of generative demand, progressing from comparative interpretation (Template 1), to exploratory association (Template 2), and finally to explicit ideation (Template 3), while sharing a common structure: each explicitly references two concepts (X and Y) without prescribing how they should be connected. This design ensured that any emergent conceptual blending arose from the generative dynamics of the model. Template indices were converted from 0–2 (Python implementation) to 1–3 for clarity.

Judge prompt

The judge module is guided by a constrained evaluation prompt that defines an explicit scoring protocol. Dream outputs (*dream_out*) are processed by the judge to (i) assign a consistency score on a discrete five-point Likert scale, and (ii) identify the most salient novel idea, if present, by summarizing in a single sentence (*idea_dream*). To enforce strict separation between evaluation and generation, the judge is required to return only a valid JSON object. This structured output enables machine-readable evaluation and makes the evaluation criteria explicit and inspectable. The extracted *idea_dream* is passed to the rewake step when the consistency score of the corresponding dream output is 4 or greater. Wake outputs (*wake_out*) are also processed by the judge to extract a single sentence (*idea_wake*) for analytical control.

Rewake prompt

The rewake stage employs a minimal articulation prompt that instructs the wake model to present a selected idea to the user within a fixed token limit (2Z+120 tokens) to avoid premature truncation. The prompt does not request additional creativity, interpretation, or expansion, but frames the task as re-articulation of an existing idea.

## 10. Sampling parameters

For each prompt, we systematically varied:

- output word limit (Z = 75, 150, 300)
- sampling temperature of the dream module (T = 0, 0.3, 0.6, 1, 3, 10)
- random seed controlling stochastic sampling (seeds = 0-4)

All other generation parameters were held constant within each sweep.

## 11. Implementation details

All experiments were conducted on an NVIDIA DGX Spark system and an Apple Mac mini (M4). Local LLM modules were executed using the llama-cpp-python bindings, which allowed controlled specification of sampling parameters, random seeds, and execution order. The wake, dream, and judge modules were instantiated using four local models: gpt-oss-120b (OSS, OpenAI), qwen3.5-27b (Qwen, Alibaba), gemma4-31b (Gemma, Google), and nemo3-30b (Nemotron, NVIDIA), designated as **o**, **q**, **g**, and **n**, respectively (Table 1). Model configurations are denoted by three-letter codes indicating the models assigned to the wake, dream, and judge modules in that order; for example, **oqg** represents OSS as wake, Qwen as dream, and Gemma as judge.

## 12. Quantifying changes in novelty, alignment, and coherence

Cliff's δ was used to quantify the change in novelty, alignment, and coherence across the wake, dream, and rewake phases. For each model configuration, δ was computed by comparing all pairwise combinations of novelty scores between two groups, yielding a measure of the overall tendency of novelty enhancement across varying parameter combinations without assuming a specific distribution [19]. For each stage comparison, scores from the later stage were specified as the first distribution and scores from the earlier stage as the second distribution (e.g. rewake minus wake). Accordingly, positive δ values indicate higher novelty at the later stage. Effect sizes were interpreted as negligible ($|\delta| < 0.147$), small ($0.147 \le |\delta| < 0.33$), medium ($0.33 \le |\delta| < 0.474$), or large ($|\delta| \ge 0.474$) [20]. The same thresholds were applied to differences in δ between each heterogeneous configuration and its corresponding homogeneous baseline in the role effectiveness index, where ++ and + indicate medium or large, and small incremental improvements, respectively (Tables 3b–6b). The mean novelty difference (Δmean) was also reported as a supplementary measure. Novelty was rated on a 5-point Likert scale by the review LLM (gpt-5.2).

## 13. Statistical Analysis

Cliff's δ was used as the primary non-parametric effect size for all pairwise novelty comparisons.

For unmatched wake–rewake comparisons (Tables 2 and 3a–6a) and wake–dream comparisons (Tables 3c–5c), the two groups were treated as independent samples, and statistical significance was assessed using the two-sided Mann–Whitney U test. For dream–rewake comparisons (Tables 3d–5d), where each rewake output is causally linked to a specific dream output via a shared run_id, the two-sided Wilcoxon signed-rank test was applied to matched pairs. Between-theme comparisons of Cliff's δ distributions were also assessed with the Wilcoxon signed-rank test (Table 7a). All tests were two-sided with a significance threshold of $p < 0.05$.

## Results

### 1. Embedding-based similarity analysis under homogeneous model conditions

To quantify semantic transformation induced by the dream phase independently of qualitative evaluation, we first performed an embedding-based similarity analysis under homogeneous model conditions, in which the same LLM was used for the wake, dream, and judge modules. Four homogeneous configurations were examined: **ooo**, **qqq**, **ggg**, and **nnn**. For each run, the wake and dream outputs (*wake_out* and *dream_out*) were embedded into a fixed-dimensional semantic space using a pretrained SentenceTransformer model, and cosine similarity was computed between the baseline wake output (*wake_out*) and the corresponding dream output (*dream_out*).

As a negative control, cosine similarity was also computed between repeated wake outputs generated under identical low-temperature conditions. These wake–wake comparisons yielded similarity values tightly concentrated near 1.0 across all homogeneous model configurations, confirming that low-temperature wake generation produces highly stable semantic representations. In contrast, wake–dream similarity values were broadly distributed over a substantially wider range in all four configurations (dream temperature = 1, 3, 10), indicating that high-temperature dream generation induces robust semantic displacement regardless of the specific LLM family used (Figure 2a, c, e, g).

To examine the effect of dream temperature on embedding diversity, we further compared the distributions of cosine similarity between *wake_out* and *dream_out* across temperature settings (Figure 2b, d, f, h). Although higher temperature settings tended to produce lower wake–dream similarity in some configurations, the overall pattern was broadly consistent across **ooo**, **qqq**, **ggg**, and **nnn**. These results suggest that temperature modulation may expand the conceptual search space, but does not deterministically control the magnitude of semantic displacement.

### 2. Quantitative evaluation of ideation quality under homogeneous model conditions

We next examined how ideational quality changes across the wake, dream, and rewake stages under homogeneous model conditions (**ooo**, **qqq**, **ggg**, and **nnn**; Table 2). The judge assigned a consistency score to each dream output, yet this score was strongly skewed toward the maximum value (5: fully consistent), indicating that obvious inconsistencies were rare. Nevertheless, pass rates fell below 50% in some conditions, because the rewake phase was skipped when the judge did not identify a novel idea in the dream output. These results suggest that the judge functions primarily as a novelty filter selectively forwarding only dream outputs that contain a novel idea, rather than as a consistency checker.

To quantitatively assess ideation quality, wake, dream, and rewake outputs were independently evaluated using an external LLM (gpt-5.2) with respect to **alignment**, **coherence**, and **novelty**. Among these metrics, novelty exhibited the most pronounced and systematic changes across processing stages. We quantified changes in **novelty** using Cliff's δ and ranked model configurations according to their novelty enhancement (Tables 3, 4, 5, 6). Among the four homogeneous conditions, **qqq** exhibited clear increases in novelty from wake to rewake outputs in Time & Space and Periodic Tarot tasks (Table 2a, b, Figure 3b, f). In contrast, **ooo** and **nnn** showed an increase in novelty in the Time & Space task (Table 2a, Figure 3a, d) but not in the Periodic Tarot task (Table 2b, Figure 3e, h). **ggg** generally failed to produce substantial novelty gains after rewake consolidation (Table 2a-c, Figure 3c, g). There was no clear increase in novelty in the Aperiodic Craft task (Table 2c).

Because **qqq** exhibited the most robust novelty enhancement while **ooo** and **nnn** showed modest improvements, Qwen, OSS, and Nemotron were selected as the baseline models for subsequent heterogeneous-model analyses.

**3. Functional specialization among LLMs in heterogeneous ReMIND configurations**

We next examined whether different LLMs were preferentially suited to specific ReMIND roles using the following fifteen heterogeneous model configurations: **qoq, qqo, qoo, gqq, qgq, qqg, qnq, qqn, oqo, ooq, oqq, nqq, noq, nqo, noo**. Nineteen model configurations, which include homogeneous and heterogeneous conditions using the Time & Space prompt, were ranked based on the changes in **novelty** between the wake and rewake phases using Cliff's δ (Table 3a, Figure 4). The results revealed substantial differences among model families depending on their assigned role within the ReMIND architecture.

We assessed role-specific model tendencies by comparing heterogeneous configurations with their corresponding homogeneous baselines, such as **qqq** and **ooo**. Compared with the **qqq** and **ooo** baselines, replacing the wake module with Nemotron, yielding **nqq** and **noo**, substantially increased novelty, indicating that Nemotron is particularly effective in the wake phase (Table 3a, b). Replacing the wake module with Qwen from **ooo**, yielding **qoo**, improved overall performance. OSS showed beneficial effects in the dream and judge roles, whereas Qwen and Nemotron did not show beneficial effects in these roles. Gemma did not show a significant effect in any of the three roles. Thus, different LLM families exhibited role-dependent functional tendencies in the ReMIND framework.

Similar role-dependent model tendencies were observed for the Periodic Tarot prompt (Table 4a, b; Figure 5), whereas overall novelty enhancement was less pronounced for Aperiodic Craft (Table 5a; Figure 6). Nevertheless, the rankings of model configurations were broadly consistent across the three prompt themes, with **nqo**, **noo**, and **noq** repeatedly appearing among the top-performing configurations, while the relative positions of other configurations varied across themes (Tables 3a–6a). Although role-specific model preference were less pronounced for Aperiodic Craft (Table 5b), the role-effectiveness analysis across all three themes showed tendencies broadly similar to those observed for Time & Space and Periodic Tarot (Table 6b). Since Cliff's δ only shows relative change in novelty between wake and rewake phases, we also considered the ranking of rewake novelty as an absolute criterion (Table 3e, 4e, 5e, 6c).

The combined ranking based on wake-to-rewake Cliff's δ and mean rewake novelty across all three themes identified **nqo**, **qoq**, and **qoo** as the top three LLM configurations (Table 6a, c, d). This result was broadly consistent with the observed suitability of Nemotron and Qwen for the wake role and OSS for the dream role, while both OSS and Qwen appeared effective in the judge role in high-performing configurations such as **nqo** and **qoq**.

## 4. Two-stage novelty enhancement through exploration and consolidation

Across multiple configurations, novelty appeared to increase from wake to dream outputs and increase further from dream to rewake outputs (Figures 4, 5). The ReMIND pipeline may enhance novelty through two complementary stages: an exploratory stage (wake→dream) and a consolidation stage (dream→rewake). However, the violin plots include dream outputs from runs that did not proceed to the rewake phase (unmatched analysis; Figures 4, 5), potentially inflating the apparent improvement in the dream phase. Notably, the matched Cliff's δ analysis, restricted to logs in which the judge successfully extracted an *idea_dream*, revealed that different LLM configurations tend to specialize in different stages (Tables 3d-5d). For example, **noo, noq, nqq,** and **nqo** showed significant improvement primarily in the wake-to-dream transition, while **qnq, qgq**, and **qoq** showed significant improvement in the dream-to-rewake transition in the Time & Space prompt (Table 3c, d).

A similar two-stage enhancement was observed for the Periodic Tarot prompt, where six configurations showed significant improvement in the wake-to-dream transition and 11 of 19 configurations achieved a significant overall novelty gain (Table 4a, c). However, stage-specific specialization in the dream-to-rewake transition was less pronounced than in the Time & Space prompt. Only **qgq** showed significant improvement at this stage (Table 4d), suggesting that the wake-to-dream transition was the primary driver of novelty enhancement for this prompt. In contrast, two-stage enhancement was unclear for the Aperiodic Craft prompt (Figure 6; Table 5a, c, d).

The first stage corresponds to exploratory generation during the dream phase. High-temperature sampling expands the conceptual search space and generates semantically distant associations that are rarely observed under low-temperature conditions. The second stage occurs during rewake consolidation. Rather than merely reformulating dream outputs, the rewake module may transform selected conceptual fragments into more explicit and interpretable ideas, resulting in additional gains in novelty.

A plausible explanation for this phenomenon lies in the interaction between the judge and wake modules. The judge module does not simply evaluate complete dream outputs; instead, it extracts a single representative sentence capturing the most salient novel concept generated during exploration. The wake module then interprets and elaborates this condensed conceptual fragment into a coherent idea during the rewake stage, as exemplified in Supplementary Text 1 and discussed further in Section 6. In this sense, the judge functions as a semantic bottleneck that isolates potentially valuable concepts, while the wake module acts as a conceptual amplifier that reconstructs and contextualizes them.

**5. Dependence of novelty enhancement on prompt themes**

To examine whether the effectiveness of ReMIND depends on the structure of the problem domain, we compared novelty enhancement across three prompt themes: Time & Space, Periodic Tarot, and Aperiodic Craft (Table 7). Substantial differences were observed among prompt themes. The strongest novelty enhancement was obtained with the Time & Space prompt, followed by the Periodic Tarot prompt. In contrast, the Aperiodic Craft prompt exhibited little or no systematic novelty enhancement across most model configurations (Table 7a).

Notably, the Aperiodic Craft prompt produced comparatively high novelty scores at the wake stage (Table 7b). Consequently, the limited novelty enhancement observed for this prompt may be attributable to a ceiling effect, as the high baseline novelty at the wake stage left limited room for further improvement during the dream phase. Wake-stage novelty was also affected by the wake model. OSS in the wake role tended to produce higher wake novelty for the Periodic Tarot and Aperiodic Craft prompts, whereas Qwen tended to produce higher wake novelty for the Aperiodic Craft prompt (Table 7c). These observations indicate that the capacity of ReMIND to enhance novelty is not uniform across tasks but is influenced by the structure of the conceptual search space and the characteristics of the LLM employed.

**6. Representative outputs generated by ReMIND**

To illustrate the qualitative characteristics of ideas generated by ReMIND, we examined representative outputs from selected configurations across the Time & Space, Periodic Tarot, and Aperiodic Craft prompt themes (Table 8; Supplementary Text).

In the **qnq** condition (run ID 177), the Time & Space prompt was processed with a dream temperature of 10.0 (Supplementary Text 1). The wake output (novelty = 3) provided a conventional description of spacetime based on relativity (Supplementary Text 1a): “Time and space are not separate, independent stages upon which the universe plays out; they are inextricably woven into a single, four-dimensional fabric known as spacetime." Most of the dream output remained within the same conceptual context, but it introduced the seemingly unrelated word "**PULLEY**" in the statement: "THIS RELATEDNESS CAN BE VISUALIZED AS A **PULLEY** THAT DRIVES A CLOCKWISE ROTATION OF GEODESICS THROUGH SPATIAL COORDINATES" (Supplementary Text 1b). The judge extracted this unusual association as the central idea: "Time acts as a **pulley** that drives a clockwise rotation of geodesics through spatial coordinates, reforming the very topology of space" (Supplementary Text 1c). During rewake, the wake module developed this extracted association into a coherent speculative account in which the pulley metaphor became the organizing concept of the final output (novelty = 5; Supplementary Text 1d). This example shows how a single conceptually distant element introduced during the dream phase can be extracted by the judge and consolidated during rewake.

In the **noo** condition (run ID 132), also from the Time & Space theme, novelty increased progressively across all three phases (2 → 4 → 5; Supplementary Text 2). The rewake output proposed a self-similar **fractal spacetime** model in which the roles of space and time vary across scales and causality becomes scale dependent. It further suggested that spacetime discreteness relevant to quantum gravity could emerge naturally from this fractal geometry.

From the Periodic Tarot theme, the **qoq** condition (run ID 797; novelty: 4 → 4 → 5; Supplementary Text 3a) produced a rewake output proposing a divinatory system inspired by stoichiometric chemistry, in which **the periodic table replaces traditional Tarot cards** (Table 8). For example, Carbon, Hydrogen, and Oxygen were associated with Glucose ($C_6H_{12}O_6$), symbolizing nourishment and growth. Once the compound is identified, **it becomes a symbolic "spell"** or ritual focus, linking chemical composition to divinatory interpretation.

A related but structurally distinct approach was produced by the **ogg** condition (run ID 336; rewake novelty = 5; Supplementary Text 3b). Rather than constructing a compound from chemical elements, this output proposed a "**Cosmic Deck**," a card deck in which all 118 chemical elements are assigned tarot-like archetypes, with periodic-table groups mapped onto traditional suit categories. This run was generated using a preliminary Gemma3-based configuration and was not included in the main quantitative analysis.

Finally, in the Aperiodic Craft theme, the **qoq** condition (run ID 531; novelty: 4 → 4 → 5; Supplementary Text 4) proposed **tactile motifs that physically enforce aperiodic matching rules** (Table 8). Rather than relying on color-coding or geometric markings, the tile edges are decorated with raised knot and stitch motifs. Incorrectly matched tiles do not fit together, whereas compatible patterns interlock correctly. This design translates the matching rules of Penrose-type tilings into tactile constraints while also using the motifs as decorative elements.

Taken together, these examples illustrate several distinct forms of idea generation within the ReMIND framework. They also show how exploratory content generated during the dream phase can be extracted by the judge and further developed by the wake module during the rewake phase.

### 7. Effects on Alignment and Coherence

The ReMIND system is designed to enhance novelty; however, our review process also evaluates alignment (responsiveness to the prompt) and coherence (internal logical consistency). A Cliff's δ ranking across all 19 configurations revealed that alignment declined substantially for certain LLMs, particularly when Gemma was used for the wake phase (Supplementary Table S1a). Coherence, by contrast, varied comparatively little across LLMs (Supplementary Table S1b). Notably, among the configurations with the largest novelty gains, coherence declined only marginally for **nqo**, **qoq**, and **qoo** (all $|\delta| < 0.147$). Alignment likewise showed no meaningful decline for **nqo** ($\delta=+0.154$, an improvement) and **qoo** ($\delta=-0.063$), although **qoq** exhibited a modest decline ($\delta=-0.197$). These metrics may nonetheless prove more critical in other tasks built on the ReMIND framework.

## Discussion

### 1. Mechanism of novelty enhancement

The present results reveal that novelty enhancement can arise through two functionally separable stages within the ReMIND pipeline. The dream phase expands the conceptual search space through high-temperature generation, while the wake module consolidates selected conceptual seeds into coherent, articulated forms during rewake. This separation parallels divergent exploration and convergent elaboration in creative problem solving, while making the two processes independently controllable within the ReMIND pipeline. Indeed, in the **qnq** 177 case, the dream phase introduced the seemingly unrelated concept of a “pulley” (Supplementary Text 1). The judge extracted this concept, and the wake module subsequently developed it during rewake into a coherent speculative model.

Heterogeneous LLM configurations often outperformed their corresponding homogeneous baselines (Tables 3-6), suggesting that individual models may exhibit different functional tendencies: some are better suited to exploratory generation during the dream phase, while others support idea selection or coherent elaboration during rewake. This finding is broadly consistent with prior multi-agent studies showing that role differentiation and interaction structure can influence collaborative ideation and problem solving [11-13].

The Role Effectiveness Table (Table 6b) shows that the contribution of a given model depends on the phase in which it is used, supporting the view that novelty enhancement can emerge from coordinated model diversity rather than solely from the capacity of a single model. ReMIND's modular design formalizes this principle: the wake, dream, and judge roles can be assigned to different models, allowing flexible substitution according to their functional suitability.

This design differs from iterative self-refinement and natural-language reflection approaches, which improve outputs through repeated feedback and revision [9,10]. It also differs from methods that structure reasoning within a single model, such as Chain-of-Thought prompting, and from architectures such as Mixture-of-Experts, which distribute computation across internal expert components [21, 22]. ReMIND instead organizes exploration, evaluation, and stabilization into sequential but functionally distinct modules to support emergent ideation.

**2. Conditions affecting novelty enhancement**

The effectiveness of ReMIND varied substantially across prompt themes, with the strongest novelty enhancement observed in the Time & Space theme, followed by Periodic Tarot, and minimal systematic enhancement in the Aperiodic Craft theme. This pattern may reflect the underlying structure of the conceptual search space associated with each prompt rather than a limitation of the pipeline per se.

Time & Space may present a broadly open conceptual domain in which the space of possible ideas is large and loosely constrained by disciplinary convention. Periodic Tarot, while constrained by the requirement to integrate two predefined conceptual domains, retains sufficient openness for creative recombination. Aperiodic Craft, by contrast, produced comparatively high novelty scores at the wake stage (Table 7b, Figure 6), suggesting that the prompt itself already invites conceptually unusual associations and leaves limited room for further enhancement through the dream phase. This possible ceiling effect suggests that ReMIND may be most effective when initial generation is anchored in familiar or conventional territory, and less suited to prompts that already elicit high-novelty associations at baseline.

The wake model additionally shaped the conceptual starting point for the pipeline. OSS-based wake configurations tended to produce higher baseline novelty in the Periodic Tarot and Aperiodic Craft themes, while Qwen-based wake configurations showed elevated wake novelty in the Aperiodic Craft theme (Table 7c). This suggests that wake model selection does not affect all prompt themes equally, and that the benefit of subsequent pipeline stages depends in part on the baseline novelty established during the wake phase.

**3. Context dependence and limits of novelty evaluation**

It remains difficult to predict which outputs will ultimately be perceived as particularly valuable, insightful, or creative. ReMIND influences novelty through model selection, role assignment, and prompt theme, but these factors alone do not fully determine the perceived value of an individual idea.

One reason is that the perception of novelty is inherently context-dependent. The perceived originality of an idea depends not only on the generated content, but also on the prior knowledge, expertise, and conceptual framework of the evaluator. An idea regarded as highly innovative by one individual may appear obvious to another who is already familiar with similar concepts. Consequently, novelty cannot be understood as a purely intrinsic property of a generated text; it emerges from the interaction between the idea and the evaluator.

Importantly, this limitation applies not only to automated evaluation but also to human evaluation. Human judgments of creativity are similarly shaped by disciplinary background, cultural context, and personal experience. While human evaluation remains valuable for assessing the practical or aesthetic significance of generated ideas, it does not provide an objective ground truth for creativity. Instead, both human and LLM-based evaluations should be regarded as context-dependent perspectives on the same underlying phenomenon.

For this reason, the present study does not attempt to establish a definitive measure of creativity. Rather, we focus on the computational processes that influence the emergence of novel ideas within the ReMIND framework. By analyzing exploration, selection, and consolidation as distinct functional stages, we seek to characterize the conditions under which serendipitous ideas are more likely to emerge, while acknowledging that the ultimate value of any particular idea may remain irreducible to quantitative metrics.

**4. Limitations and future directions**

Several limitations of the present study warrant consideration. First, novelty evaluation relied on a single LLM-based scoring system, which raises a potential concern about circularity: outputs generated by LLMs were also evaluated by an LLM. LLM-based evaluators provide scalable and reproducible assessment of open-ended outputs, but they are also known to exhibit biases related to model preference, response style, and evaluation design [16,17]. Although the evaluation model was held constant across conditions to enable fair comparison, the absolute novelty scores obtained may not generalise to other evaluative contexts or to human judgement. Second, the present analyses were conducted across three prompt themes, which constitute a limited sample of possible creative tasks. The generalisability of the observed phenomena remains to be established across a broader set of prompt themes. Third, all configurations were evaluated under fixed generation parameters per condition; systematic exploration of the interaction between model identity, temperature, and prompt structure across a wider parameter space was beyond the scope of the present study.

A natural direction for future work is fine-tuning LLMs for the specific functional demands of the dream and judge roles. In the present study, general-purpose models were deployed across all phases. However, the dream phase imposes distinctive requirements that may not be optimally served by models trained for instruction-following or factual accuracy. Similarly, the judge role requires reliable discrimination between genuinely novel conceptual configurations and superficially unusual but substantively trivial outputs, a capacity that dedicated training data and reward signals could improve. Fine-tuning models for these roles represents a concrete path toward improving both the yield and the quality of ReMIND-generated ideas.

Beyond model optimization, the pipeline architecture itself admits several extensions. The present implementation treats the wake–dream–rewake sequence as a single pass. Iterative extensions, such as running multiple dream cycles within a single pass and selecting the most productive output, or using rewake outputs as wake inputs for subsequent cycles, may enable more extensive exploration and progressive refinement of emerging ideas. Recent studies have demonstrated the feasibility of using single- and multi-agent LLM systems for scientific idea generation [2,12,13,23]. Extending ReMIND to more structured or goal-directed tasks, including scientific hypothesis generation, design exploration, and analogical reasoning, would test the generality of the modular orchestration principles proposed here.

**Acknowledgements**

This work was supported by Grant-in-Aid for Scientific Research (A) and (B), Grant-in-Aid for Transformative Research Areas (A) from MEXT (22H05169, 22H05621, 24H00188, 24H01396, and 25K02282 to M.S.).

**Author contributions**

M.S. conceived and designed the study, implemented the code, performed the computational experiments, analyzed the data, interpreted the results, and wrote the manuscript.

**Data availability statement**

The custom code developed in this study is available online (https://github.com/satouma7/ReMIND).

**Conflicts of Interest**

The authors declare no conflicts of interest associated with this manuscript.

## Figures

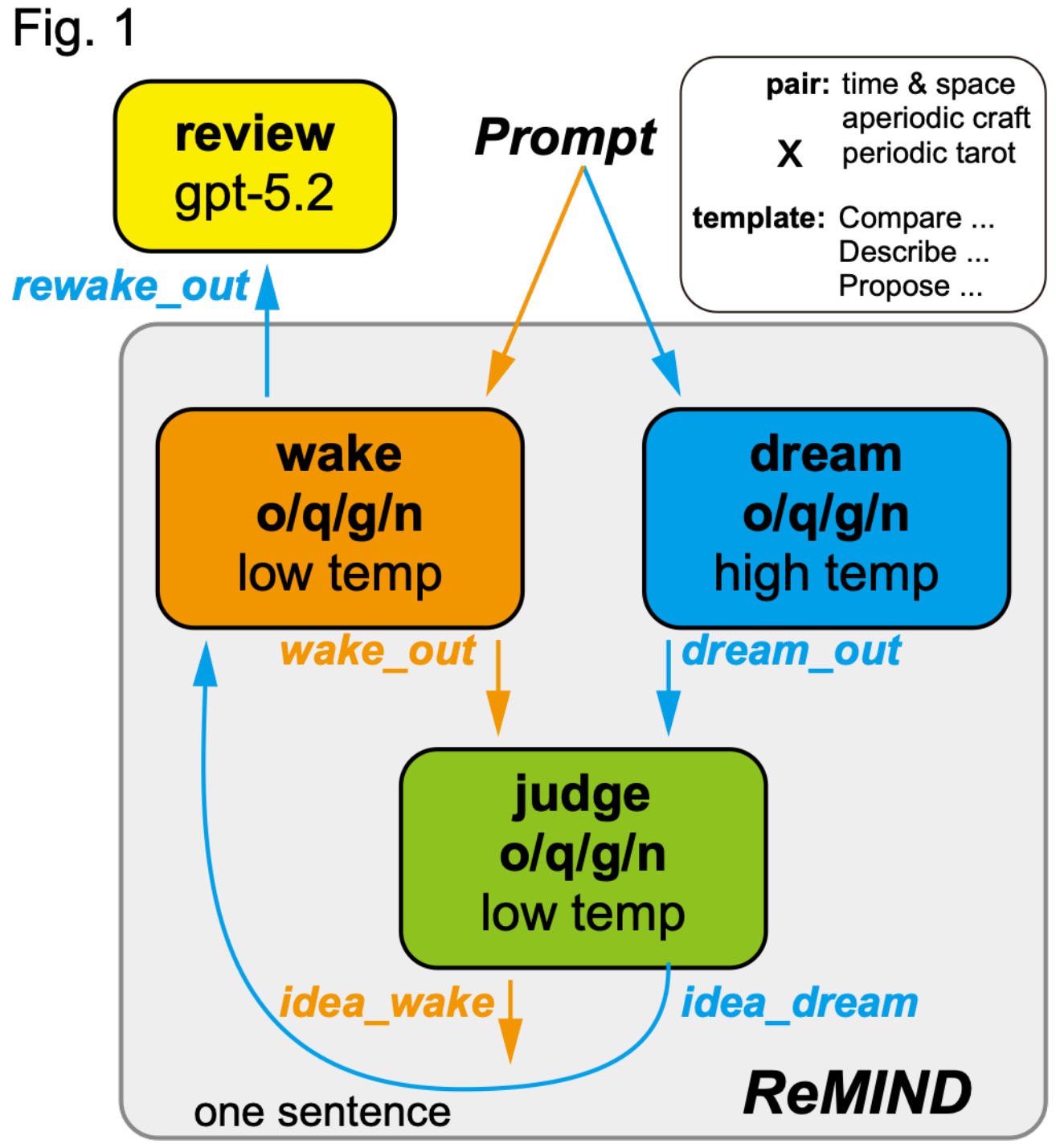

**Figure 1. Schematic overview of the ReMIND framework.** ReMIND orchestrates three functionally distinct LLM modules in sequence. Model configurations are denoted by three-letter codes corresponding to the wake, dream, and judge modules, respectively: **o**, gpt-oss120b; **q**, qwen3.5_27b; **g**, gemma4_31n; and **n**, nemo3_30b. The wake module generates a low-temperature baseline response to the input prompt, establishing a stable semantic anchor (*wake_out*). The dream module performs high-temperature stochastic generation, producing exploratory outputs that deviate from the baseline (*dream_out)*. The judge module evaluates the wake and dream outputs for consistency and extracts the most salient novel concept as a single sentence (*idea_wake* and *idea_dream*). *idea_dream* is passed back to the wake module, which re-articulates the extracted concept into a coherent final output (*rewake_out*). Three outputs (*wake_out, dream_out, rewake_out*) are evaluated by the review LLM (gpt-5.2).

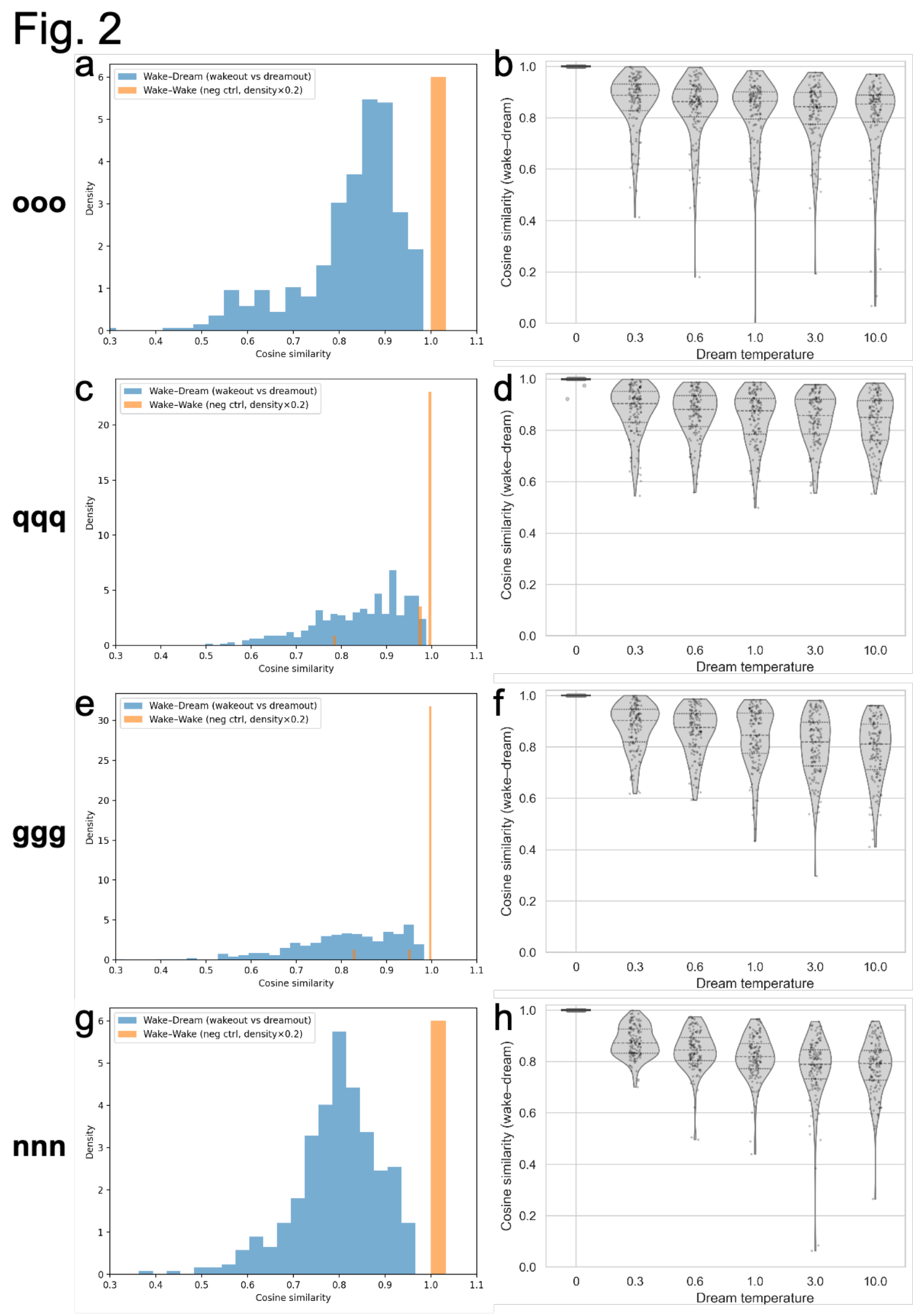

**Figure 2. Embedding-based cosine similarity analysis under homogeneous model conditions.** Distributions of cosine similarity between wake and dream outputs (wake–dream) and between wake outputs (wake–wake) are shown for four homogeneous configurations: **ooo** (a, b), **qqq** (c, d), **ggg** (e, f), and **nnn** (g, h). **a, c, e, g,** Histogram distributions of wake–wake (negative control) and wake–dream cosine similarity (dream temperature = 1, 3, 10), computed using a pretrained SentenceTransformer model. Wake–wake similarity values are tightly concentrated near 1.0, whereas wake–dream similarity is broadly distributed at lower values, indicating robust semantic displacement induced by high-temperature dream generation. Runs with dream temperature below 1.0 were excluded. **b, d, f, h,** Distributions of wake–dream cosine similarity stratified by dream temperature (0, 0.3, 0.6, 1, 3, 10). Higher temperatures tended to produce lower similarity values.

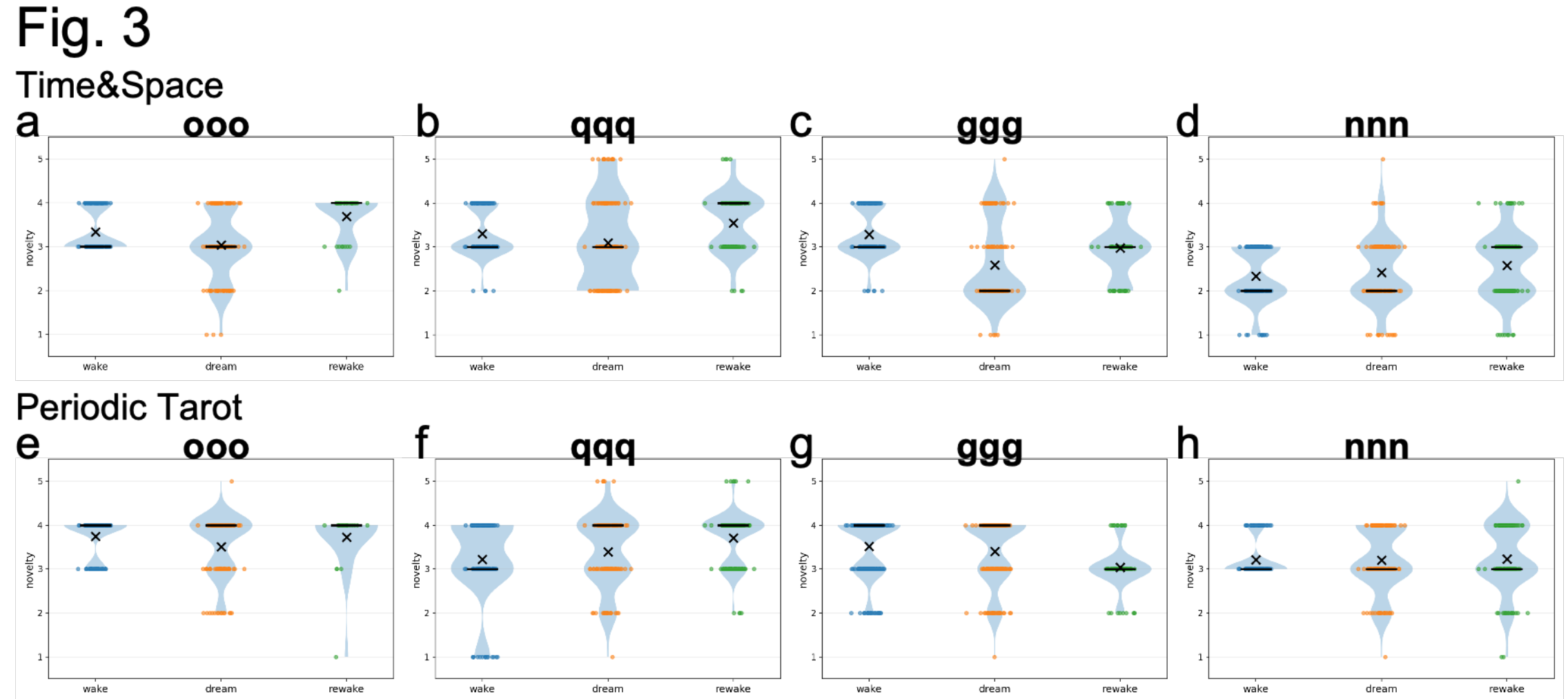

**Figure 3. Novelty distributions across pipeline stages under homogeneous model conditions.** Violin plots show the distribution of novelty scores at the wake, dream, and rewake stages for four homogeneous configurations under the Time & Space (a–d) and Periodic Tarot (e–h) prompts. Panels correspond to **ooo** (a, e), **qqq** (b, f), **ggg** (c, g), and **nnn** (d, h). The Aperiodic Craft prompt is not shown, as no systematic novelty enhancement was observed. Novelty was scored on a five-point Likert scale (1–5) by an independent review LLM (gpt-5.2). Dream outputs that were discarded by the judge are included in the analysis (unmatched). These conventions apply to Figures 3–7.

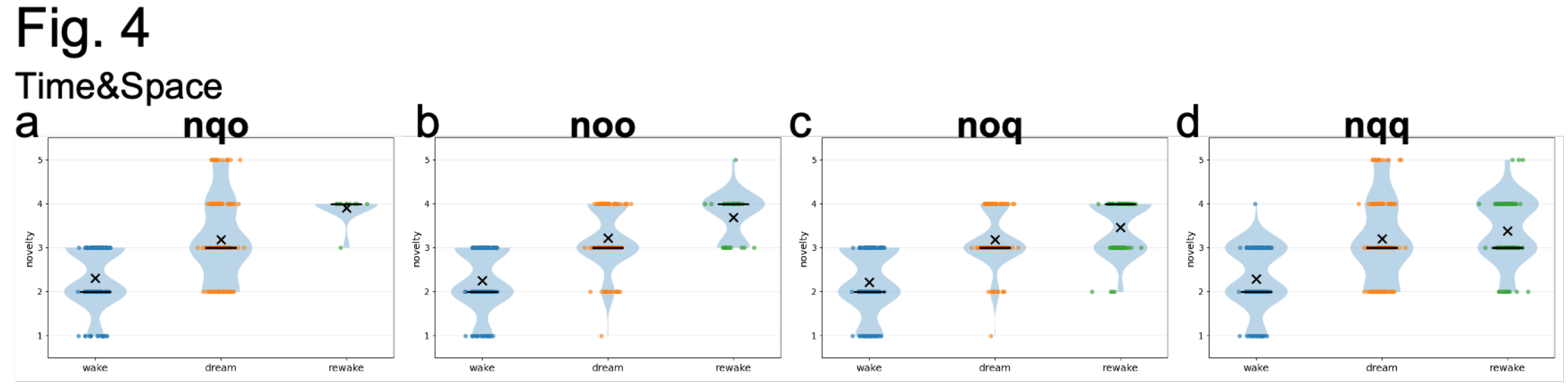

**Figure 4. Novelty distributions for the top-performing configurations in the Time & Space prompt.** Violin plots show the distribution of novelty scores at the wake, dream, and rewake stages for the four highest-ranked configurations under the Time & Space prompt, based on Cliff's δ: **nqo** (a), **noo** (b), **noq** (c), and **nqq** (d). Scoring and analysis conventions as in Figure 3.

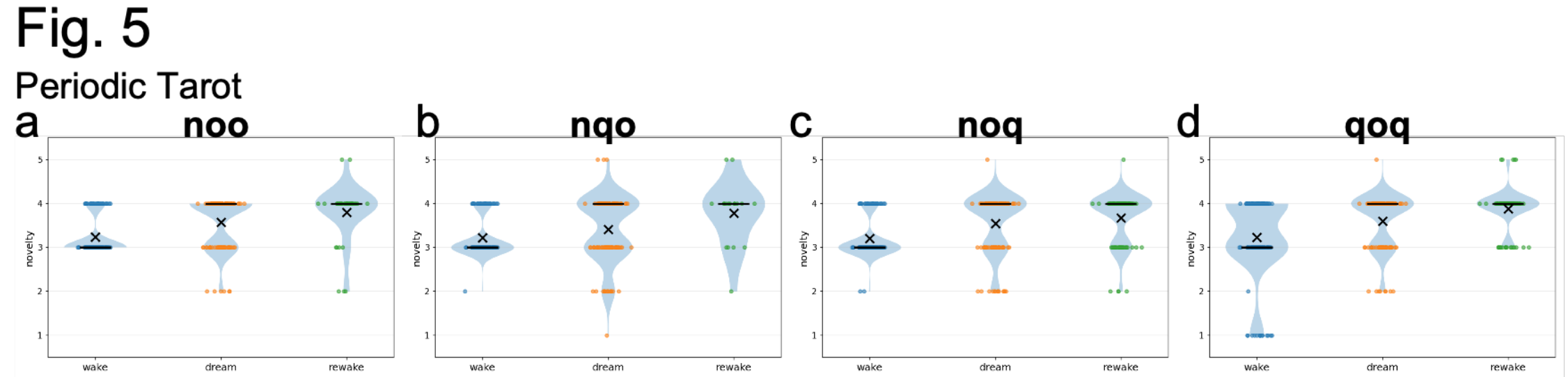

**Figure 5. Novelty distributions for the top-performing configurations in the Periodic Tarot prompt.** Violin plots show the distribution of novelty scores at the wake, dream, and rewake stages for the four highest-ranked configurations under the Periodic Tarot prompt, based on Cliff's δ: **noo** (a), **nqo** (b), **noq** (c), and **qoq** (d). Scoring and analysis conventions as in Figure 3.

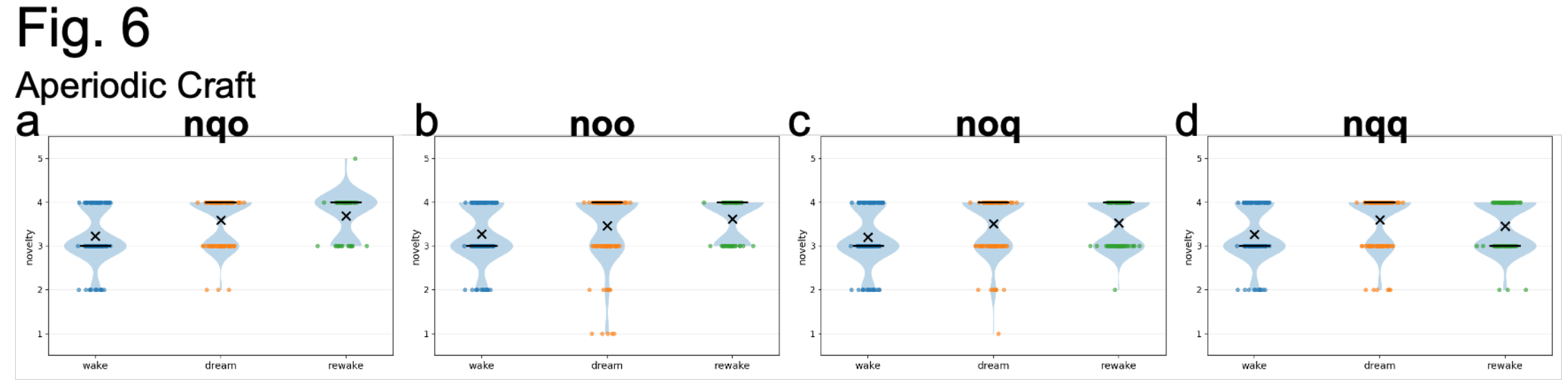

**Figure 6. Novelty distributions for the top-performing configurations in the Aperiodic Craft prompt.** Violin plots show the distribution of novelty scores at the wake, dream, and rewake stages for the four highest-ranked configurations under the Aperiodic Craft prompt, based on Cliff's δ: **nqo** (a), **noo** (b), **noq** (c), and **nqq** (d). Scoring and analysis conventions as in Figure 3.

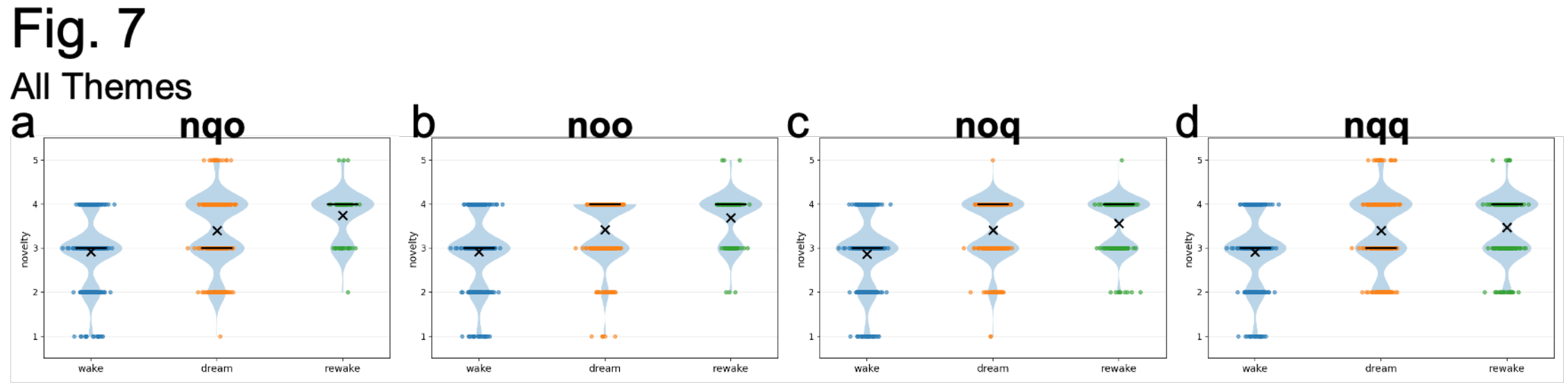

**Figure 7. Novelty distributions for the top-performing configurations across all prompt themes.** Violin plots show the distribution of novelty scores at the wake, dream, and rewake stages for the four highest-ranked configurations in the all-themes analysis, based on Cliff's δ: **nqo** (a), **noo** (b), **noq** (c), and **nqq** (d). Scoring and analysis conventions as in Figure 3.

**Tables**

| Abbreviation | Model name | GGUF filename | Developer |
|---|---|---|---|
| o | gpt-oss120b | gpt-oss-120b-MXFP4-00001-of-00002.gguf | OpenAI |
| q | qwen35_27b | Qwen3.5-27B-Q4_K_M.gguf | Alibaba |
| g | gemma4_31b | gemma-4-31B-it-Q4_K_M.gguf | Google |
| n | nemo3_30b | Nemotron-3-Nano-30B-A3B-Q4_K_M.gguf | Nvidia |

**Table 1. LLM models used in the ReMIND pipeline.** Model abbreviations correspond to the single-letter codes used throughout the text. GGUF filenames indicate the quantised models used for local inference via llama-cpp-python.

a. Time&Space

| rank | cond | Cliff's δ | Δmean | p | n | pass rate | judge mean | judge median |
|---|---|---|---|---|---|---|---|---|
| 1 | **ooo** | 0.357 | 0.347 | <0.001 | 32 | 24% | 4.393 | 5 |
| 2 | **qqq** | 0.215 | 0.246 | 0.0019 | 91 | 67% | 4.985 | 5 |
| 3 | **nnn** | 0.172 | 0.25 | 0.0076 | 132 | 98% | 4.896 | 5 |
| 4 | ggg | -0.246 | -0.315 | <0.001 | 77 | 57% | 4.437 | 5 |

b. Periodic Tarot

| rank | cond | Cliff's δ | Δmean | p | n | pass rate | judge mean | judge median |
|---|---|---|---|---|---|---|---|---|
| 1 | **qqq** | 0.268 | 0.46 | <0.001 | 85 | 63% | 4.867 | 5 |
| 2 | nnn | 0.059 | -0.021 | 0.5558 | 22 | 16% | 4.556 | 5 |
| 3 | ooo | 0.051 | 0.012 | 0.3946 | 132 | 98% | 4.8 | 5 |
| 4 | ggg | -0.233 | -0.26 | 0.0012 | 83 | 61% | 4.822 | 5 |

c. Aperiodic Craft

| rank | cond | Cliff's δ | Δmean | p | n | pass rate | judge mean | judge median |
|---|---|---|---|---|---|---|---|---|
| 1 | nnn | 0.003 | -0.02 | 0.9583 | 128 | 95% | 4.859 | 5 |
| 2 | ooo | -0.028 | -0.028 | 0.7177 | 57 | 42% | 4.828 | 5 |
| 3 | qqq | -0.085 | -0.085 | 0.1958 | 101 | 75% | 4.919 | 5 |
| 4 | ggg | -0.341 | -0.43 | <0.001 | 90 | 67% | 4.667 | 5 |

**Table 2. Novelty enhancement statistics for homogeneous model configurations.** Summary of novelty change from wake to rewake phases for four homogeneous configurations (**ooo, qqq, ggg, nnn**) across three prompt themes: Time & Space (a), Periodic Tarot (b), and Aperiodic Craft (c). Cliff's δ and mean novelty difference (Δmean) quantify the magnitude of novelty change; *p*-values are based on Mann–Whitney U tests (unmatched). *n* indicates the number of rewake outputs. *pass rate* indicates the success rate to obtain the rewake outputs. *judge mean* and *judge median* are the mean and median of the consistency score evaluated by the judge.

a. Time&Space ranking

| rank | cond | Cliff's δ | Δmean | p | n |
|---|---|---|---|---|---|
| **1** | **nqo** | 0.961 | 1.598 | <0.0001 | 11 |
| **2** | **noo** | 0.867 | 1.443 | <0.0001 | 36 |
| **3** | **noq** | 0.765 | 1.249 | <0.0001 | 82 |
| **4** | **nqq** | 0.663 | 1.092 | <0.0001 | 92 |
| **5** | **qoo** | 0.510 | 0.535 | <0.0001 | 36 |
| **6** | **oqo** | 0.449 | 0.428 | 0.0018 | 13 |
| **7** | **qoq** | 0.434 | 0.454 | <0.0001 | 76 |
| **8** | **qqo** | 0.422 | 0.439 | 0.0180 | 8 |
| **9** | **ooo** | 0.357 | 0.347 | 0.0002 | 32 |
| **10** | **ooq** | 0.300 | 0.320 | <0.0001 | 79 |
| **11** | **qqq** | 0.215 | 0.246 | 0.0019 | 91 |
| **12** | **qqg** | 0.181 | 0.170 | 0.0105 | 86 |
| **13** | **nnn** | 0.172 | 0.250 | 0.0076 | 132 |
| 14 | oqq | 0.130 | 0.101 | 0.0560 | 92 |
| 15 | qgq | 0.112 | 0.123 | 0.1099 | 77 |
| 16 | gqq | 0.040 | 0.020 | 0.5575 | 92 |
| 17 | qnq | -0.051 | -0.079 | 0.4575 | 86 |
| 18 | qqn | -0.056 | -0.081 | 0.3607 | 135 |
| 19 | ggg | -0.246 | -0.315 | 0.0006 | 77 |

b. Role effectiveness

| LLM | Wake | Dream | Judge |
|---|---|---|---|
| o | +/− | + | + |
| q | + | +/− | +/− |
| g | +/− | +/− | +/− |
| n | ++ | − | − |

c. Wake-Dream novelty change

| rank | cond | Cliff's δ | Δmean | p | n |
|---|---|---|---|---|---|
| **1** | **noo** | 0.634 | 0.9700 | <0.0001 | 135 |
| **2** | **noq** | 0.633 | 0.970 | <0.0001 | 135 |
| **3** | **nqq** | 0.501 | 0.911 | <0.0001 | 135 |
| **4** | **nqo** | 0.488 | 0.874 | <0.0001 | 135 |
| 5 | nnn | 0.036 | 0.081 | 0.5661 | 135 |
|  | others | <0 |  |  |  |

d. Dream-Rewake novelty change

| rank | cond | Cliff's δ | Δmean | p | n |
|---|---|---|---|---|---|
| **1** | **qnq** | 0.434 | 0.616 | <0.0001 | 86 |
| **2** | **qgq** | 0.245 | 0.403 | <0.0001 | 77 |
| **3** | **qoq** | 0.181 | 0.237 | 0.0038 | 76 |
| **4** | **nnn** | 0.118 | 0.159 | 0.0087 | 132 |
| 5 | qqn | 0.100 | 0.096 | 0.0851 | 135 |
| 6 | ooq | 0.088 | 0.114 | 0.0606 | 79 |
| 7 | qqq | 0.085 | 0.154 | 0.0534 | 91 |
| 8 | ggg | 0.062 | 0.065 | 0.4821 | 77 |

e. Rewake ranking

| rank | cond | rewake | wake | n |
|---|---|---|---|---|
| 1 | nqo | 3.909 | 2.311 | 11 |
| 2 | qoo | 3.861 | 3.326 | 36 |
| 3 | oqo | 3.769 | 3.341 | 13 |
| 4 | qoq | 3.750 | 3.296 | 76 |
| 5 | qqo | 3.750 | 3.311 | 8 |
| 6 | noo | 3.694 | 2.252 | 36 |
| 7 | ooo | 3.688 | 3.341 | 32 |
| 8 | ooq | 3.646 | 3.326 | 79 |
| 9 | qqq | 3.549 | 3.304 | 91 |
| 10 | qqg | 3.488 | 3.319 | 86 |
| 11 | noq | 3.463 | 2.215 | 82 |
| 12 | qgq | 3.442 | 3.319 | 77 |
| 13 | oqq | 3.435 | 3.333 | 92 |
| 14 | nqq | 3.380 | 2.289 | 92 |
| 15 | gqq | 3.272 | 3.252 | 92 |
| 16 | qnq | 3.233 | 3.311 | 86 |
| 17 | qqn | 3.200 | 3.281 | 135 |
| 18 | ggg | 2.974 | 3.289 | 77 |
| 19 | nnn | 2.583 | 2.333 | 132 |

**Table 3. Novelty enhancement analysis for the Time & Space prompt.** (a) Cliff's δ ranking of all 19 model configurations for overall novelty change from wake to rewake. (b) Role effectiveness table showing the effect of substituting each LLM into the wake, dream, or judge role relative to the corresponding homogeneous baseline. (c) Stage-specific analysis of the wake-to-dream transition: Cliff's δ, Δmean, *p*-value (Mann–Whitney U test; wake and dream outputs are generated independently from the same prompt), and sample size (*n*). (d) Stage-specific analysis of the dream-to-rewake transition: Cliff's δ, Δmean, *p*-value (Wilcoxon signed-rank test; dream and rewake outputs are matched by run ID), and sample size (*n*). (e) Ranking of model configurations by mean rewake novelty score. Statistical notation is consistent across Tables 3–5.

a. Periodic Tarot ranking

| rank | cond | Cliff's δ | Δmean | p | n |
|---|---|---|---|---|---|
| **1** | **noo** | 0.543 | 0.563 | <0.0001 | 40 |
| **2** | **nqo** | 0.477 | 0.556 | <0.0001 | 18 |
| **3** | **noq** | 0.467 | 0.477 | <0.0001 | 96 |
| **4** | **qoq** | 0.429 | 0.653 | <0.0001 | 102 |
| **5** | **qqo** | 0.428 | 0.596 | 0.0038 | 14 |
| **6** | **qoo** | 0.425 | 0.631 | <0.0001 | 40 |
| **7** | **nqq** | 0.375 | 0.407 | <0.0001 | 85 |
| **8** | **qqq** | 0.268 | 0.46 | 0.0002 | 85 |
| **9** | **qgq** | 0.203 | 0.372 | 0.0036 | 92 |
| **10** | **qqg** | 0.181 | 0.332 | 0.0172 | 73 |
| **11** | **ooq** | 0.151 | 0.151 | 0.0116 | 97 |
| 12 | qnq | 0.09 | 0.243 | 0.1916 | 98 |
| 13 | oqq | 0.081 | 0.054 | 0.2151 | 85 |
| 14 | oqo | 0.067 | 0.042 | 0.4884 | 22 |
| 15 | ooo | 0.059 | -0.021 | 0.5558 | 22 |
| 16 | nnn | 0.051 | 0.012 | 0.3946 | 132 |
| 17 | qqn | -0.01 | 0.06 | 0.8739 | 131 |
| 18 | gqq | -0.054 | -0.064 | 0.4472 | 85 |
| 19 | ggg | -0.233 | -0.26 | 0.0012 | 83 |

b. Role effectiveness

| LLM | Wake | Dream | Judge |
|---|---|---|---|
| o | − | + | + |
| q | ++ | +/− | +/− |
| g | − | +/− | +/− |
| n | +/− | − | − |

c. Wake-Dream novelty change

| rank | cond | Cliff's δ | Δmean | p | n |
|---|---|---|---|---|---|
| **1** | **noq** | 0.354 | 0.3480 | <0.0001 | 135 |
| **2** | **noo** | 0.344 | 0.333 | <0.0001 | 135 |
| **3** | **qoq** | 0.223 | 0.370 | 0.0004 | 135 |
| **4** | **nqq** | 0.208 | 0.222 | 0.0009 | 135 |
| **5** | **nqo** | 0.205 | 0.193 | 0.0009 | 135 |
| **6** | **qoo** | 0.193 | 0.348 | 0.0022 | 135 |
| 15 | qgq | -0.072 | -0.022 | 0.2617 | 135 |

d. Dream-Rewake novelty change

| rank | cond | Cliff's δ | Δmean | p | n |
|---|---|---|---|---|---|
| **1** | **qgq** | 0.188 | 0.2390 | 0.0018 | 92 |
| 2 | qoq | 0.033 | 0.039 | 0.4142 | 102 |
| 3~18 | others | <0 | | n.s. | |

e. Rewake ranking

| rank | cond | rewake | wake | n |
|---|---|---|---|---|
| 1 | qoq | 3.882 | 3.230 | 102 |
| 2 | qoo | 3.875 | 3.244 | 40 |
| 3 | qqo | 3.857 | 3.261 | 14 |
| 4 | ooq | 3.825 | 3.674 | 97 |
| 5 | oqo | 3.818 | 3.776 | 22 |
| 6 | noo | 3.800 | 3.237 | 40 |
| 7 | nqo | 3.778 | 3.222 | 18 |
| 8 | ooo | 3.727 | 3.748 | 22 |
| 9 | oqq | 3.706 | 3.652 | 85 |
| 10 | qqq | 3.682 | 3.222 | 85 |
| 11 | noq | 3.677 | 3.200 | 96 |
| 12 | qgq | 3.609 | 3.237 | 92 |
| 13 | nqq | 3.600 | 3.193 | 85 |
| 14 | qqg | 3.562 | 3.230 | 73 |
| 15 | qnq | 3.480 | 3.237 | 98 |
| 16 | gqq | 3.306 | 3.370 | 85 |
| 17 | qqn | 3.290 | 3.230 | 131 |
| 18 | nnn | 3.227 | 3.215 | 132 |
| 19 | ggg | 3.036 | 3.296 | 83 |

**Table 4. Novelty enhancement analysis for the Periodic Tarot prompt.** (a) Cliff's δ ranking of all 19 model configurations for overall novelty change from wake to rewake. (b) Role effectiveness table showing the effect of substituting each LLM into the wake, dream, or judge role relative to the corresponding homogeneous baseline. (c) Stage-specific analysis of the wake-to-dream transition. (d) Stage-specific analysis of the dream-to-rewake transition. (e) Ranking of model configurations by mean rewake novelty score.

a. Aperiodic Craft ranking

| rank | cond | Cliff's δ | Δmean | p | n |
|---|---|---|---|---|---|
| 1 | **nqo** | 0.386 | 0.466 | <0.0001 | 46 |
| **2** | **noo** | 0.282 | 0.347 | 0.0005 | 58 |
| **3** | **noq** | 0.267 | 0.332 | 0.0001 | 94 |
| **4** | **nqq** | 0.156 | 0.189 | 0.0206 | 101 |
| **5** | **qoq** | 0.145 | 0.154 | 0.0157 | 95 |
| 6 | qoo | 0.054 | 0.065 | 0.4594 | 58 |
| 7 | oqo | 0.017 | 0.017 | 0.8276 | 51 |
| 8 | ooq | 0.011 | 0.004 | 0.8674 | 94 |
| 9 | nnn | 0.003 | -0.02 | 0.9583 | 128 |
| 10 | ooo | -0.028 | -0.028 | 0.7177 | 57 |
| 11 | qqo | -0.052 | -0.052 | 0.5526 | 37 |
| 12 | oqq | -0.079 | -0.092 | 0.2119 | 101 |
| 13 | qqq | -0.085 | -0.085 | 0.1958 | 101 |
| 14 | qqg | -0.108 | -0.103 | 0.1025 | 93 |
| 15 | qnq | -0.153 | -0.165 | 0.0208 | 100 |
| 16 | qgq | -0.164 | -0.182 | 0.0268 | 63 |
| 17 | qqn | -0.17 | -0.194 | 0.0056 | 133 |
| 18 | gqq | -0.196 | -0.246 | 0.0045 | 101 |

b. Role effectiveness

| LLM | Wake | Dream | Judge |
|---|---|---|---|
| o | +/− | + | +/− |
| q | +/− | +/− | +/− |
| g | − | +/− | +/− |
| n | + | +/− | +/− |

c. Wake-Dream novelty change

| rank | cond | Cliff's δ | Δmean | p | n |
|---|---|---|---|---|---|
| **1** | **nqo** | 0.311 | 0.3630 | <0.0001 | 135 |
| **2** | **nqq** | 0.292 | 0.333 | <0.0001 | 135 |
| **3** | **noq** | 0.264 | 0.311 | <0.0001 | 135 |
| **4** | **noo** | 0.196 | 0.193 | 0.0020 | 135 |
| 5 | gqq | 0.118 | 0.163 | 0.0570 | 135 |
| 6 | nnn | 0.029 | -0.007 | 0.6486 | 135 |
| | others | <0 | | | |

d. Dream-Rewake novelty change

| rank | cond | Cliff's δ | Δmean | p | n |
|---|---|---|---|---|---|
| **1** | **qoq** | -0.002 | 0.0000 | 1.0000 | 95 |
| | others | <0 | | | |

e. Rewake ranking

| rank | cond | rewake | wake | n |
|---|---|---|---|---|
| 1 | qoq | 3.821 | 3.667 | 95 |
| 2 | qoo | 3.724 | 3.659 | 58 |
| 3 | nqo | 3.696 | 3.230 | 46 |
| 4 | ooq | 3.670 | 3.667 | 94 |
| 5 | oqo | 3.647 | 3.630 | 51 |
| 6 | qqo | 3.622 | 3.674 | 37 |
| 7 | noo | 3.621 | 3.274 | 58 |
| 8 | ooo | 3.614 | 3.642 | 57 |
| 9 | oqq | 3.574 | 3.667 | 101 |
| 10 | qqg | 3.548 | 3.652 | 93 |
| 11 | noq | 3.532 | 3.200 | 94 |
| 12 | qqq | 3.515 | 3.600 | 101 |
| 13 | qgq | 3.492 | 3.674 | 63 |
| 14 | nqq | 3.455 | 3.267 | 101 |
| 15 | qnq | 3.450 | 3.615 | 100 |
| 16 | qqn | 3.429 | 3.622 | 133 |
| 17 | nnn | 3.195 | 3.215 | 128 |
| 18 | gqq | 3.168 | 3.415 | 101 |

**Table 5. Novelty enhancement analysis for the Aperiodic Craft prompt.** (a) Cliff's δ ranking of all 19 model configurations for overall novelty change from wake to rewake. (b) Role effectiveness table showing the effect of substituting each LLM into the wake, dream, or judge role relative to the corresponding homogeneous baseline. (c) Stage-specific analysis of the wake-to-dream transition. (d) Stage-specific analysis of the dream-to-rewake transition. (e) Ranking of model configurations by mean rewake novelty score.

a. All Theme ranking

| rank | cond | Cliff's δ | Δmean | p | n |
|---|---|---|---|---|---|
| **1** | **nqo** | 0.595 | 0.826 | <0.0001 | 75 |
| **2** | **noo** | 0.549 | 0.773 | <0.0001 | 134 |
| **3** | **noq** | 0.488 | 0.691 | <0.0001 | 272 |
| **4** | **nqq** | 0.388 | 0.559 | <0.0001 | 278 |
| **5** | **qoq** | 0.343 | 0.427 | <0.0001 | 273 |
| **6** | **qoo** | 0.321 | 0.396 | <0.0001 | 134 |
| **7** | **qqo** | 0.225 | 0.279 | 0.0016 | 59 |
| **8** | **ooq** | 0.157 | 0.163 | <0.0001 | 270 |
| 9 | qqq | 0.135 | 0.202 | 0.0007 | 277 |
| 10 | oqo | 0.135 | 0.128 | 0.0209 | 86 |
| 11 | ooo | 0.100 | 0.081 | 0.0566 | 111 |
| 12 | qqg | 0.090 | 0.132 | 0.0288 | 252 |
| 13 | nnn | 0.059 | 0.079 | 0.1146 | 392 |
| 14 | qgq | 0.058 | 0.112 | 0.1628 | 232 |
| 15 | oqq | 0.044 | 0.018 | 0.2608 | 278 |
| 16 | qnq | -0.024 | 0.007 | 0.5381 | 284 |
| 17 | gqq | -0.073 | -0.101 | 0.0707 | 278 |
| 18 | qqn | -0.076 | -0.072 | 0.0366 | 399 |
| 19 | ggg | -0.275 | -0.335 | <0.0001 | 250 |

b. Role effectiveness

| LLM | Wake | Dream | Judge |
|---|---|---|---|
| o | +/– | + | +/- |
| q | + | +/- | +/- |
| g | − | +/- | +/- |
| n | + | - | - |

d. Total ranking

| rank | cond | Cliff's δ | rewake |
|---|---|---|---|
| **1** | **nqo** | 1 | 3 |
| **2** | **qoq** | 5 | 1 |
| **3** | **qoo** | 6 | 2 |
| 4 | noo | 2 | 7 |
| 5 | noq | 3 | 11 |

c. Rewake ranking

| rank | cond | rewake | wake | n |
|---|---|---|---|---|
| 1 | qoq | 3.824 | 3.398 | 273 |
| 2 | qoo | 3.806 | 3.410 | 134 |
| 3 | nqo | 3.747 | 2.921 | 75 |
| 4 | ooq | 3.719 | 3.556 | 270 |
| 5 | oqo | 3.709 | 3.582 | 86 |
| 6 | qqo | 3.695 | 3.416 | 59 |
| 7 | noo | 3.694 | 2.921 | 134 |
| 8 | ooo | 3.658 | 3.577 | 111 |
| 9 | qqq | 3.578 | 3.375 | 277 |
| 10 | oqq | 3.568 | 3.551 | 278 |
| 11 | noq | 3.562 | 2.872 | 272 |
| 12 | qqg | 3.532 | 3.400 | 252 |
| 13 | qgq | 3.522 | 3.410 | 232 |
| 14 | nqq | 3.475 | 2.916 | 278 |
| 15 | qnq | 3.394 | 3.388 | 284 |
| 16 | qqn | 3.306 | 3.378 | 399 |
| 17 | gqq | 3.245 | 3.346 | 278 |
| 18 | nnn | 3.000 | 2.921 | 392 |
| 19 | ggg | 2.988 | 3.323 | 250 |

**Table 6. Novelty enhancement analysis across all prompt themes.** (a) Cliff's δ ranking of all 19 model configurations for overall novelty change from wake to rewake, averaged across themes. (b) Role effectiveness table showing the effect of substituting each LLM into the wake, dream, or judge role, averaged across themes. (c) Ranking of model configurations by mean rewake novelty score across themes. (d) Total ranking based on the combined sum of Cliff's δ rank and rewake novelty rank.

a. Difference between prompt pairs

| Theme | mean Cliff's δ | mean Δmean | p vs periodic | p vs aperiodic |
|---|---|---|---|---|
| time & space | 0.328 | 0.439 | 0.026 * | <0.0001 *** |
| periodic tarot | 0.210 | 0.277 | - | <0.0001 *** |
| aperiodic craft | -0.003 | -0.001 | - | - |

b. Wake novelty (different Theme)

| Theme | mean | median | n |
|---|---|---|---|
| time & space | 3.039 | 3 | 13 / 19 |
| periodic tarot | 3.341 | 3 | 11 / 19 |
| **aperiodic craft** | 3.513 | 4 | 5 / 19 |

c. Wake novelty (diferent Theme and Wake model)

| Theme | o-wake | q-wake | g-wake | n-wake |
|---|---|---|---|---|
| time & space | 3.335 | 3.308 | 3.270 | 2.280 |
| periodic tarot | 3.712 | 3.236 | 3.333 | 3.213 |
| aperiodic craft | 3.651 | 3.645 | 3.400 | 3.237 |
| mean | 3.566 | 3.397 | 3.335 | 2.910 |

**Table 7. Influence of prompt theme on novelty enhancement.** (a) Pairwise comparison of wake-rewake novelty change across prompt themes (Time & Space, Periodic Tarot, Aperiodic Craft); mean Cliff's δ and mean novelty difference (Δmean) are shown for each theme. Pairwise *p*-values were computed using the two-sided Wilcoxon signed-rank test on the 19 per-condition Cliff's δ values matched across themes (19 pairs per comparison). (b) Wake-stage novelty by theme: mean, median, and the number of model configurations showing significant novelty improvement (Cliff's $\delta < 0$ and $p < 0.05$) out of 19 conditions. (c) Mean wake novelty stratified by prompt theme and wake model (n-wake, o-wake, q-wake, g-wake).

| # | Theme | cond | run_id | template | word | temp | seed | novelty (W→D→R) |
|---|---|---|---|---|---|---|---|---|
| 1 | time & space | qnq | 177 | 1 | 300 | 10 | 1 | 3 → 3 → 5 |
| 2 | time & space | noo | 132 | 2 | 300 | 10 | 1 | 2 → 4 → 5 |
| 3a | periodic tarot | qoq | 797 | 2 | 300 | 1 | 1 | 4 → 4 → 5 |
| 3b | periodic tarot | ogg | 336 | 1 | 150 | 3 | 0 | R: 5 |
| 4 | aperiodic craft | qoq | 531 | 2 | 300 | 3 | 0 | 4 → 4 → 5 |

**Table 8. Parameters and novelty scores for representative ReMIND outputs.** Generation parameters and novelty scores for the five example runs. Novelty scores are reported for the wake (W), dream (D), and rewake (R) stages as evaluated by the review LLM (gpt-5.2). Wake and dream novelty scores are unavailable for run 3b, which was generated under a preliminary configuration outside the main evaluation pipeline (Gemma3 was used rather than Gemma4).

**Supplementary Text**

**<u>1. qnq</u>** <u>177 (Time & Space)</u>

**a. wake**

Time and space are not separate, independent stages upon which the universe plays out; they are inextricably woven into a single, four-dimensional fabric known as spacetime. This unexpected unity means that one cannot exist without the other, and their measurements are relative to the observer's motion.

As an object moves faster through space, it simultaneously slows its progression through time. This phenomenon, called time dilation, reveals that velocity trades spatial distance for temporal duration. For a traveler moving near the speed of light, years might pass on Earth while only days elapse for them. Space and time effectively exchange roles depending on perspective; what one observer measures as purely a difference in location, another might perceive as a mix of space and time intervals.

Furthermore, mass and energy do not merely sit within this fabric; they actively warp it. Gravity is not a mysterious force pulling objects together but rather the curvature of spacetime itself. A massive object like Earth creates a dip in this four-dimensional sheet, causing nearby objects to follow curved paths that we interpret as gravitational attraction. In this view, falling toward a planet is simply moving along the straightest possible line through warped geometry.

The relationship becomes even stranger near black holes, where extreme curvature can stretch time to a standstill relative to distant observers while compressing space into a singularity. Here, the distinction between "where" and "when" blurs completely. The future direction of time for an infalling object points inevitably toward the center, just as spatial directions point outward elsewhere.

Ultimately, the universe does not consist of three dimensions of space evolving over one dimension of time. Instead, it is a static block where past, present, and future coexist in a complex geometric structure. Our perception of flowing time is merely our consciousness traversing this unified manifold. Space and time are two sides of the same coin, dynamically interacting to define the very reality we inhabit.

**b. dream**

TIME IS A HIDDEN GEOMETRICAL CURVATURE THAT MOLDS SPACE. WHILE SPACE ORDINARYLY CONJURES AN INFINITE THREE-DIMENSIONAL GRID OF POINTS, TIME BENDS THAT GRID LIKE AN INVISIBLE ARCHWAY THAT STRETCHES FROM EVERY DIRECTION INTO A FOURTH DIMENSION. THE INTERACTION BETWEEN TIME'S LINEAR ADVANCEMENT AND SPACE'S THREE-AXIAL EXTENT PRODUCES A SYNERGISTIC ENVELOPE WHERE CAUSAL SEQUENCES MAP ONTO EVERY PHYSICAL COORDINATE, SHAPING WHAT CAN BE PERCEIVED AS REALITY.

THIS INTERDEPENDENCE IS UNEXPECTED BECAUSE TIME IS USUALLY THOUGHT OF AS AN ORDINARY SERIAL FLOW, YET WHEN INTEGRATED INTO THE METRIC THAT GOVERNS SPACE-TIME IT TRANSFORMS FROM A SIMPLE BACKGROUND TO AN ACTIVE CURVATURE. THE RESULTING ENTITY—THE COSMIC MANIFOLD—

ENCAPSULATES EVERY PHENOMENON FROM STELLAR EVOLUTION TO QUANTUM FLUCTUATIONS.

MORE SURPRISINGLY, THIS RELATEDNESS CAN BE VISUALIZED AS A **PULLEY** THAT DRIVES A CLOCKWISE ROTATION OF GEODESICS THROUGH SPATIAL COORDINATES. TIME'S FLOW DEFINES THE DIRECTION OF MOTION IN THE HIGHEST-DIMENSIONAL ARRAY, MAKING EACH STEP NOT JUST A MOVE BUT AN ESSENTIAL SHIFT THAT REFORMS THE VERY TOPOLOGY OF SPACE.

THUS, THE UNEXPECTED RELATIONSHIP IS THAT TIME IS NOT AN EXTERNAL TICK-TACKER ON A STATIC CANVAS; IT IS THE INTEGRATIVE FORCE THAT CARRIES AND CONTINUES SPACE'S SHAPE THROUGH CONTINUOUS INTERWEAVING, CREATING A DYNAMIC ARCHITECTURE WHERE TIME CONSTANTLY RECONSTRUCTS SPACE WHILE SPACE PROVIDES THE STAGE FOR TIME'S ENDLESS ADVANCE.

**c. idea_dream**

Time acts as a **pulley** that drives a clockwise rotation of geodesics through spatial coordinates, reforming the very topology of space.

**d. rewake**

Imagine a universe where time is not merely a passive backdrop but an active mechanical force. Consider this radical proposition: Time acts as a **pulley** that drives a clockwise rotation of geodesics through spatial coordinates, fundamentally reforming the very topology of space.

In standard physics, geodesics represent the straightest possible paths in curved spacetime, dictated by mass and energy. However, under this new paradigm, time itself becomes the tension on a cosmic cable. As it pulls, it imparts a specific rotational momentum to these paths. This is not a simple spin of matter within space; rather, the trajectories themselves twist clockwise as they traverse dimensions.

This continuous rotation alters how points in space connect to one another. Just as twisting a rope changes its structural integrity and shape, this temporal **pulley** reshapes the fabric of reality. Regions that were once distant might become adjacent through topological folding, while familiar geometries could unravel into complex, non-Euclidean structures. The "straight line" between two stars is no longer static; it is a dynamic spiral driven by the relentless pull of time.

This mechanism suggests that the expansion of the universe or the behavior of black holes might be secondary effects of this primary rotational drive. If time pulls geodesics clockwise, it could explain anomalous observations in galactic rotation curves without invoking dark matter, as the paths of light and gravity are inherently curved by the temporal **pulley** itself.

Furthermore, this implies a directional asymmetry to the cosmos rooted in the mechanics of time. The topology of space is not fixed but is constantly being knitted and re-knitted by this rotational force. We do not just move through space; we ride the currents of a twisting geometry that time actively sculpts.

Adopting this view transforms our understanding of causality and structure. It posits that the universe's shape is a direct consequence of time's mechanical action, turning the abstract concept of temporal flow into a tangible architect of spatial reality. This idea invites us to look at the cosmos not as a static stage, but as a dynamic machine where time turns the gears of existence.

**2. noo** 132 (Time & Space)

**rewake**

Imagine a universe where the familiar separation between “where” and “when” dissolves into a single, self-similar geometry—**a fractal metric field** that encodes both distance and duration as interchangeable layers of scale. At each level of magnification, coordinates are no longer fixed axes; instead they morph, swapping their roles as you move deeper or shallower into the hierarchy.

At coarse resolution, the field behaves like ordinary spacetime: a smooth manifold where points are linked by timelike or spacelike intervals. Zoom in, however, and the metric reveals intricate branching patterns reminiscent of Mandelbrot sets. Here, a step that was once a simple spatial displacement can become a temporal advance, and vice-versa, depending on the scale you choose to observe.

Mathematically, each point carries a multiscale coordinate pair $(x_s, t_s)$ where the superscript s denotes the current scaling factor. As s increases, the effective “length” of an interval contracts while its “duration” expands, preserving a conserved fractal invariant $I = \int dx \cdot dt$ across all scales. This invariance guarantees that physical laws retain their form under renormalization, but the interpretation of cause and effect can invert when crossing scale boundaries.

Consequences ripple outward: causality becomes scale-conditional; events that appear simultaneous at one level may be ordered at another. Information flow can be encoded in the branching topology itself, allowing a signal to travel through space while simultaneously propagating backward in time at finer scales, and forward again at coarser ones. The field thus supports phenomena such as “temporal echoes” where past configurations re-emerge at higher magnifications, and “spatial ghosts” where distant locations share identical metric signatures.

In practical terms, **a fractal spacetime** offers a unified language for quantum gravity, where discreteness emerges naturally from the self-similar structure. It predicts observable signatures—anomalous dispersion relations, scale-dependent speed of light, and subtle correlations in cosmic microwave background fluctuations—that could be searched for with next-generation astrophysical instruments.

Ultimately, viewing time and space as two faces of a single fractal metric invites us to rethink reality as a dynamic tapestry whose threads intertwine across scales, reshaping our intuition about motion, duration, and the very fabric that holds the cosmos together.

**3. qoq** 797 (Periodic Tarot)

**a. rewake**

Imagine a unique divinatory system where **the periodic table replaces traditional Tarot cards**. In this practice, you draw three chemical elements at random to form your reading. Each element represents an archetypal force: its atomic number dictates the intensity of the energy, while its group and period reveal the nature of the challenge or opportunity in your life. For instance, drawing Sodium (Na), Chlorine (Cl), and Oxygen (O) might suggest a volatile situation requiring purification and stability.

The true magic lies in the resolution phase. Instead of interpreting the cards statically, you must balance these three elements into a single, stable stoichiometric compound. This process mirrors the alchemical journey of integrating conflicting aspects of your psyche or circumstances. If your draw yields Carbon (C), Hydrogen (H), and Oxygen (O), you might synthesize Glucose ($C_6H_{12}O_6$) to symbolize nourishment, growth, and the sweet results of hard work. Alternatively, balancing Iron (Fe), Sulfur (S), and Oxygen (O) could yield Hematite ($Fe_2O_3$), representing grounding, protection, and resilience against external pressures.

Once the compound is identified, **it becomes your symbolic spell**. The balanced equation serves as a mantra or ritual focus. You visualize the chaotic individual elements reacting to form this new, harmonious substance, effectively "transmuting" your current struggles into a defined outcome. This method bridges the gap between scientific precision and mystical intuition, offering a structured yet creative path to insight. By grounding abstract divination in the concrete laws of chemistry, you gain a tangible framework for understanding how disparate parts of your life can combine to create something greater than the sum of its elements. Embrace this fusion of science and spirit to unlock new dimensions of self-discovery and intentional manifestation.

**b. rewake** (**ogg** 336 v1) **g**: Gemma3-27b

Imagine a “**Cosmic Deck**” that fuses the periodic table’s scientific rigor with tarot’s archetypal symbolism. Each of the 118 chemical elements becomes a uniquely illustrated card, pairing its atomic number, symbol, and key properties with a corresponding Major or Minor Arcana theme (e.g., Hydrogen — The Fool’s boundless potential; Iron — Strength/Justice). The deck’s suits mirror the table’s groups—alkali metals as Wands, halogens as Swords, noble gases as Pentacles, transition metals as Cups—while the numbered cards reflect periods and electron shells. Users can “read” the deck to explore hidden patterns: scientific insights (reactivity trends, periodic families) surface alongside intuitive messages about growth, balance, or transformation. By laying out a spread, you simultaneously glimpse elemental chemistry and archetypal meaning, revealing surprising connections between matter’s building blocks and life’s symbolic narratives.

**4. qoq** 531 (Aperiodic Craft)

**rewake**

Imagine a new approach to tiling where the rules of assembly are hidden within the art itself. Instead of relying on complex geometric instructions or color-coding, we propose decorating tile edges with **tactile motifs**—specifically knots and stitches. These raised textures serve a dual purpose: they act as beautiful visual elements while simultaneously functioning as mechanical constraints **that enforce aperiodic matching rules**.

In this system, tiles can only fit together when their specific knot patterns interlock or stitch lines align perfectly. If a user attempts to place a tile incorrectly, the physical mismatch prevents it from sitting flush with its neighbors. This tactile feedback guides the assembler intuitively toward the correct configuration without requiring any prior knowledge of Penrose tilings, quasicrystals, or advanced mathematics. The "knots" might represent specific rotational symmetries, while "stitches" could dictate edge continuity, ensuring that the resulting pattern never repeats periodically.

This method democratizes the creation of complex aperiodic structures. Architects, artists, and hobbyists can generate mathematically profound designs simply by feeling for the right fit, much like solving a puzzle where the solution is physically enforced. The aesthetic result is a surface rich with texture and depth, where the functional mechanism of the tiling becomes an integral part of the decoration. By merging form and function so seamlessly, we transform abstract mathematical constraints into tangible, sensory experiences. This approach not only simplifies the construction process but also invites a deeper appreciation for the hidden order within seemingly chaotic patterns, proving that rigorous mathematical rules can be communicated through touch as effectively as through sight.

## Supplementary Table

a. Alignment

| rank | cond | time & space | periodic tarot | aperiodic craft | mean |
|---|---|---|---|---|---|
| **1** | **ggg** | -0.637 | -0.458 | -0.802 | **-0.633** |
| **2** | **gqq** | -0.568 | -0.473 | -0.818 | **-0.620** |
| **3** | **qnq** | -0.425 | -0.426 | -0.516 | **-0.456** |
| **4** | **oqq** | -0.247 | -0.380 | -0.468 | **-0.365** |
| **5** | **nnn** | -0.156 | -0.444 | -0.484 | **-0.362** |
| **6** | **qqn** | -0.307 | -0.221 | -0.548 | **-0.359** |
| **7** | **qqq** | -0.365 | -0.126 | -0.559 | **-0.350** |
| **8** | **qgq** | -0.215 | -0.118 | -0.603 | **-0.312** |
| **9** | **qqg** | -0.152 | -0.312 | -0.448 | **-0.304** |
| **10** | **oqo** | -0.121 | -0.236 | -0.351 | **-0.236** |
| **11** | **goq** | -0.131 | 0.044 | -0.505 | **-0.197** |
| **12** | **ooq** | 0.025 | -0.198 | -0.405 | **-0.193** |
| 13 | qqo | -0.034 | -0.022 | -0.335 | -0.131 |
| 14 | ooo | 0.097 | -0.031 | -0.346 | -0.093 |
| 15 | nqq | 0.315 | -0.262 | -0.330 | -0.092 |
| 16 | goo | 0.069 | 0.057 | -0.313 | -0.063 |
| 17 | noq | 0.369 | -0.180 | -0.220 | -0.011 |
| 18 | noo | 0.464 | -0.104 | 0.000 | 0.120 |
| 19 | nqo | 0.570 | -0.004 | -0.106 | 0.154 |

b. Coherence

| rank | cond | time & space | periodic tarot | aperiodic craft | mean |
|---|---|---|---|---|---|
| **1** | **gqq** | -0.301 | -0.195 | -0.604 | **-0.366** |
| **2** | **ggg** | -0.369 | -0.055 | -0.460 | **-0.295** |
| **3** | **qnq** | -0.222 | -0.367 | -0.250 | **-0.280** |
| **4** | **qqq** | -0.175 | -0.282 | -0.257 | **-0.238** |
| **5** | **qqg** | -0.055 | -0.356 | -0.269 | **-0.227** |
| **6** | **oqq** | -0.242 | -0.212 | -0.198 | **-0.217** |
| **7** | **nqq** | -0.109 | -0.209 | -0.270 | **-0.196** |
| **8** | **qqn** | -0.052 | -0.229 | -0.278 | **-0.186** |
| 9 | ooq | -0.038 | -0.211 | -0.149 | -0.133 |
| 10 | oqo | -0.289 | -0.092 | 0.000 | -0.127 |
| 11 | nnn | 0.051 | -0.188 | -0.223 | -0.120 |
| 12 | goq | -0.030 | -0.147 | -0.158 | -0.112 |
| 13 | qgq | -0.050 | -0.065 | -0.206 | -0.107 |
| 14 | goo | -0.144 | -0.075 | -0.052 | -0.090 |
| 15 | noq | -0.026 | -0.116 | -0.125 | -0.089 |
| 16 | noo | -0.072 | -0.147 | -0.039 | -0.086 |
| 17 | nqo | 0.036 | -0.144 | -0.006 | -0.038 |
| 18 | ooo | -0.084 | 0.040 | -0.070 | -0.038 |
| 19 | qqo | 0.119 | 0.000 | 0.000 | 0.040 |

**Table S1. Alignment and Coherence analysis across all prompt themes.** Cliff's δ ranking (Wake→Rewake, unmatched) of all 19 model configurations for the change in alignment (a) and coherence (b), averaged across the three prompt themes.